\documentclass{article} 
\usepackage{collas2025_conference,times}
\usepackage{easyReview}
\usepackage{algpseudocode}
\usepackage{algorithm}
\usepackage{amsmath}

\usepackage{amsmath,amsfonts,bm}

\def\eqref#1{equation~\ref{#1}}

\def\1{\bm{1}}

\DeclareMathAlphabet{\mathsfit}{\encodingdefault}{\sfdefault}{m}{sl}
\SetMathAlphabet{\mathsfit}{bold}{\encodingdefault}{\sfdefault}{bx}{n}

\newcommand{\bx}{\mathbf{x}}
\newcommand{\bX}{\mathbf{X}}
\newcommand{\bg}{\mathbf{g}}

\newcommand{\bz}{\mathbf{z}}
\usepackage{hyperref}
\hypersetup{
    colorlinks=true,
    linkcolor=red,
    filecolor=magenta,
    urlcolor=blue,
    citecolor=purple,
    pdftitle={What can grokking teach us about learning under nonstationarity?},
    pdfpagemode=FullScreen,
    }
\usepackage{wrapfig}

\title{What Can Grokking Teach Us About Learning\\ Under Nonstationarity?}

\author{Clare Lyle, Ghada Sokar, Razvan Pascanu, András Gy\"orgy\thanks{ Correspondence to \texttt{clarelyle@google.com}.} \\
Google DeepMind\\
}

\collasfinalcopy
\begin{document}

\maketitle

\begin{abstract}
In continual learning problems, it is often necessary to overwrite components of a neural network's learned representation in response to changes in the data stream; however, neural networks often exhibit \textit{primacy bias}, whereby early training data hinders the network's ability to generalize on later tasks. 
While feature-learning dynamics of nonstationary learning problems are not well studied, the emergence of feature-learning dynamics is known to drive the phenomenon of \textit{grokking}, wherein neural networks initially memorize their training data and only later exhibit perfect generalization. This work conjectures that the same feature-learning dynamics which facilitate generalization in grokking also underlie the ability to overwrite previous \textit{learned} features as well, and methods which accelerate grokking by facilitating feature-learning dynamics are promising candidates for addressing primacy bias in non-stationary learning problems. We then propose a straightforward method to induce feature-learning dynamics as needed throughout training by increasing the \textit{effective} learning rate, i.e. the ratio between parameter and update norms.
We show that this approach both facilitates feature-learning and improves generalization in a variety of settings, including grokking, warm-starting neural network training, and reinforcement learning tasks.
\end{abstract}

\section{Introduction}

Non-stationarity is ubiquitous in real-world applications of AI systems: datasets may grow over time, correlations may appear and then disappear as trends evolve, and AI systems themselves may take an active role in the generation of their own training data. 
However, neural network training algorithms typically assume a fixed data distribution, and successfully training a neural network in non-stationary conditions requires avoiding a variety of potential failure modes, including loss of plasticity~\citep{lyle2021understanding, dohare2021continual, abbas2023loss}, where the network becomes less able to reduce its training loss over time, and primacy bias ~\citep{ash2020warm, nikishin2022primacy}, where early data impede the network's ability generalize well on later tasks\footnote{{The sensitivity of neural networks to early training data has been identified in the literature under a variety of names, for example \textit{critical periods}~\citep{achille2017critical}. We use the term \textit{primacy bias} in this work to refer to all such phenomena.}}. While many prior works have analyzed the learning dynamics that lead to loss of plasticity, comparatively less attention has been paid to the learning dynamics underlying degraded generalization ability. Current approaches to address the problem include perturbation or complete re-initialization of the network parameters~\citep{ash2020warm, schwarzer2023bigger, lee2023plastic} or regularization in some form towards the initialization distribution~\citep{lyle2021understanding, kumar2023maintaining} or initial conditions \citep{lewandowski2024learning}, a solution which, while effective, can transiently reduce performance and does not provide insight into \textit{why} the network fails to generalize in the first place.

In this paper, we will propose a framework for understanding and mitigating this degradation in generalization performance which connects three previously disparate phenomena: primacy bias, grokking, and feature-learning dynamics.
\begin{itemize}
    \item \textbf{Primacy bias:}  A neural network initially trained on one task is trained on a different data distribution and/or objective, and achieves worse performance than a randomly initialized network on the new task \citep{achille2017critical, ash2020warm, nikishin2022primacy}.

\item \textbf{Grokking:} 
    A model suddenly closes the generalization gap as a result of (possibly prolonged) further training after it has initially achieved perfect training accuracy (memorization) with poor test-time performance \citep{power2022grokking}.

\item \textbf{Feature learning:} a network's ability to make nontrivial changes to its learned representation (a.k.a. rich dynamics), defined in contrast to the lazy regime where dynamics mimic those of a linear model with respect to some fixed kernel~\citep{jacot2018neural}.
\end{itemize}

{Concretely, we investigate the hypothesis 
that the \textit{same fundamental process} by which a network replaces randomly initialized \textit{memorizing} features with \textit{generalizing} ones during grokking can be leveraged in continual learning problems to overwrite previously-learned features with new ones. Thus, a method which accelerates grokking (e.g., in modular arithmetic problems where this phenomenon is typically examined) will be expected to also mitigate primacy bias in non-stationary learning problems.}
    \begin{wrapfigure}{r}{0.58\textwidth}
    \centering
 {{\includegraphics[width=0.52\textwidth]{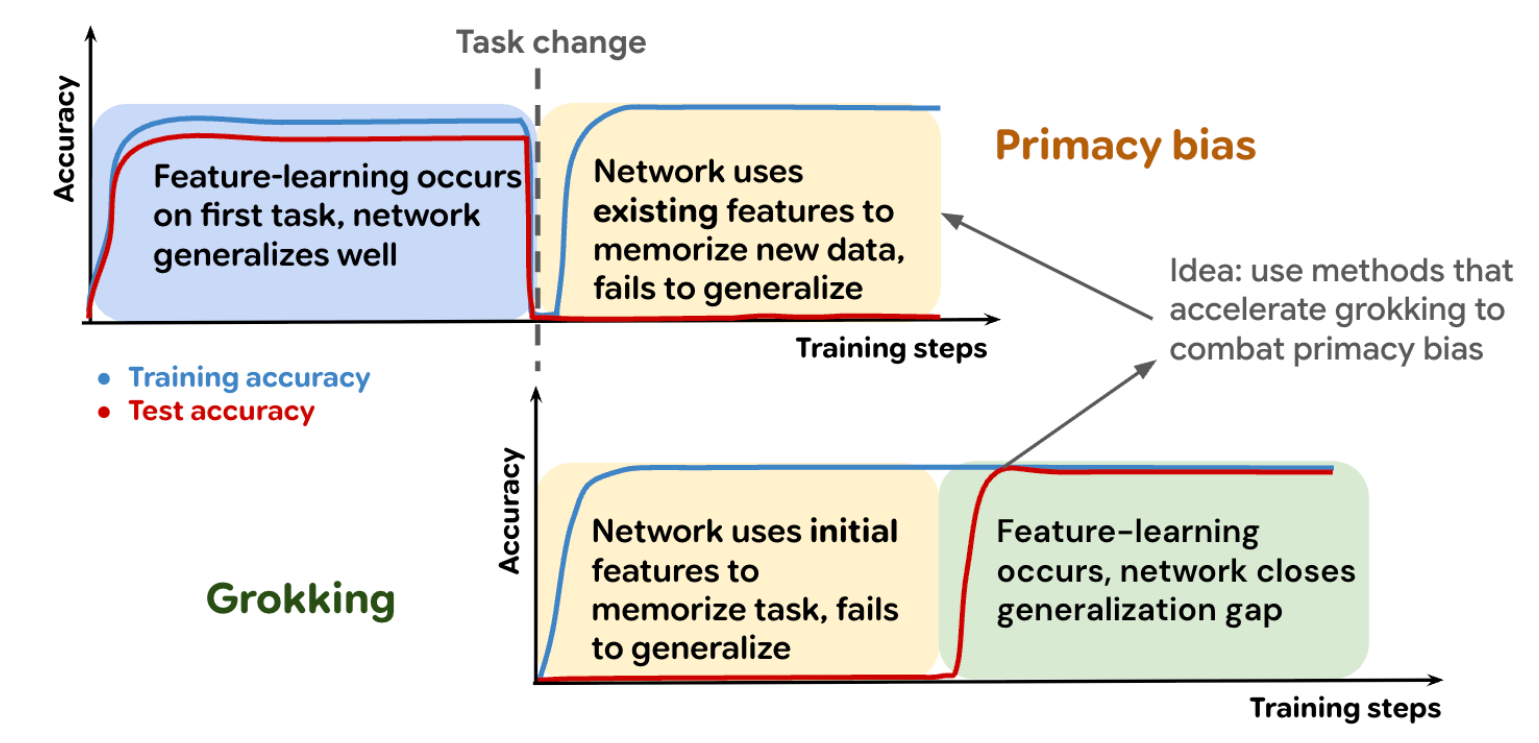}
    \caption{Both primacy bias (top row) and grokking (bottom row) exhibit a period where the network exhibits poor generalization due to an absence of feature-learning (yellow). In primacy bias, this is due to unsuitable \textit{learned} features from the initial training phase (blue). In grokking, the network eventually recovers feature-learning dynamics (green) and generalizes.}
    \label{fig:intro-figure}}}
\end{wrapfigure}
We visualize this analogy in Figure~\ref{fig:intro-figure}, which illustrates how the memorization and (possible) feature-learning phase of grokking can be mapped onto the continual learning setting. We go on to make three primary contributions.

First, based on the above hypothesis, we develop a \textbf{unifying framework} for understanding \textbf{grokking} and \textbf{primacy bias} as the \textbf{emergence (or lack thereof) of feature-learning dynamics} (Section~\ref{sec:feature-learning}).

Second, to translate the above framework into an algorithm, we propose a simple modification to the Normalize and Project method of \citet{lyle2024normalization}, which addresses the loss of plasticity problem by avoiding the excessive decay of the effective learning rate (ELR) -- a scale-invariant interpretation of the learning rate which takes into account the magnitude of the parameter norms, as defined in Section~\ref{sec:background} -- during the training process. We show that increasing the effective learning rate after it has decayed, a process which we will refer to as \textbf{ELR re-warming}, \textbf{induces feature-learning dynamics}, and propose a simple method (Algorithm~\ref{alg:cap} in Section~\ref{sec:fl-metrics}) that allows the ELR to be re-warmed periodically throughout training, recovering feature-learning dynamics outside of the initial learning period.

Finally, we successfully \textbf{apply ELR re-warming to improve generalization} on both stationary and non-stationary domains. We consider three primary applications: \textbf{grokking} (Section~\ref{sec:grokking}), where the dataset is fixed and generalization requires over-writing the features provided by the random initialization; \textbf{warm-starting} image classification (Section~\ref{sec:warm-starting}), where the network is trained on a dataset which grows over time as more data is added; and finally \textbf{reinforcement learning} (Section~\ref{sec:reinforcement-learning}), where not only the input distribution, but also the relationship between input and prediction target itself, is in flux. The success of our method in these different domains demonstrates that ELR re-warming does not depend on any special structure of the features which might arise during initialization or learning features for a given task.

Taken in the context of prior work, this paper presents a radically simplified model for primacy bias and its mitigation which is corroborated empirically in a diverse set of domains. Our analysis also reveals a trade-off analogous to that between stability and plasticity: increasing the effective learning rate allows the network shake off the influence of irrelevant early training data, but this increase must be transient to allow the optimization process to make fine-grained modifications to the network outputs necessary for optimal performance. Our findings thus present a promising springboard for further work developing the ELR as a means of navigating the landscape of plasticity-stability tradeoffs in continual learning. 

\section{Background and Related Work}
\label{sec:background}
This work provides a unifying perspective on feature learning in neural networks, connecting the disparate phenomena of \textit{plasticity loss}, the failure to adapt features to new learning signals, and \textit{grokking}, the acquisition of feature-learning dynamics after interpolation has been achieved. This section provides background on each of these topics and introduces the concept of an \textit{effective} learning rate, which will play a critical role in our proposed method.

\subsection{Loss of Plasticity in Deep Learning}

Training a neural network on a sequence of tasks has been demonstrated to interfere with its ability to adapt to new information, a phenomenon which we will refer to as \textit{loss of plasticity}~\citep{dohare2021continual,lyle2021understanding,nikishin2022primacy,dohare2024loss}. Loss of plasticity has been demonstrated to impair performance in a number of RL tasks~\citep{igl2021transient, lyle2023understanding, nikishin2023deep}, as well as in non-stationary supervised learning  problems~\citep{ash2020warm, berariu2021study}. Plasticity loss can take the form of an inability to minimize the training objective in response to changes in the data distribution, occurring in sequential learning of unrelated tasks \citep{lyle2021understanding}, or an inability to generalize to new, unseen inputs, as occurs in warm-starting \citep{lee2024slow}. A variety of approaches to mitigate plasticity loss have been proposed, such as regularization towards the initial parameters~\citep{kumar2023maintaining}, the use of normalization layers~\citep{lyle2024normalization}, weight clipping~\citep{elsayedweight}, perturbation or partial resetting of the network parameters~\citep{dohare2023loss}, and spectral regularization~\citep{lewandowski2024learning}. Loss of plasticity has been further tied both to reductions in the network's effective step size as a result of parameter norm growth~\citep{lyle2024disentangling} as well as changes in the network's curvature~\citep{lewandowski2023curvature}, quantities which have exhibited deep connections in stationary supervised learning settings via the \textit{catapult mechanism}~\citep{lewkowycz2020large}, implicit regularization~\citep{barrett2020implicit}, and edge-of-stability dynamics~\citep{cohen2021gradient, agarwala2022second, roulet2023interplay}. We will show in this work that not only does avoiding excessive decay in the effective learning rate mitigate plasticity loss, but that re-warming it to a sufficient value can recover many of the beneficial feature-learning conditions observed in the large-learning-rate early-training regime by prior works.

\subsection{Grokking}

Grokking comprises part of a broader body of work studying delayed generalization~\citep{belkin2019reconciling, heckel2021early, davies2023unifying}. While first prominently observed in small transformers trained on modular arithmetic~\citep{liu2022towards}, grokking has since been identified in a range of tasks, including parity-learning and image classification variants~\citep{baustiste2024unveiling, liu2022omnigrok}. Grokking has been theoretically characterized as a phase transition from ``lazy'' to ``rich'' learning dynamics. In the initial lazy phase, the network fits the training data without fundamentally changing its internal representations, akin to a linear model~\citep{jacot2018neural, yang2019wide}. The subsequent, much slower, rich-learning phase involves a significant reorganization of the network's learned features into a more structured, generalizing solution~\citep{xu2024benign,kumar2024grokking, lyu2024dichotomy}. Theoretical analysis of this transition is complemented by empirical observations of the structured features learned in networks which grok~\citep{nanda2023progress, liu2022towards}. In modular arithmetic tasks, for instance, progress metrics can identify the emergence of these generalizing features prior to the uptick in test performance, sometimes tracking the formation of partially-generalizing solutions along the way~\citep{nanda2023progress, varma2023explaining}.

A variety of works have drawn empirical connections between the parameter norm and grokking, noting, for example, that phase transitions in the model's generalization performance accompany periods of instability in the network's optimization dynamics and rapid increases in the parameter norm~\citep{thilak2022the}. While there is a widely observed correlation between the parameter norm and grokking, the causal relationship between weight norm and delayed generalization remains unclear. Whereas \citet{liu2022omnigrok} characterize grokking as the convergence of the parameter norm towards a ``goldilocks zone"~\citep{fort2019goldilocks} which allows for generalization, \citet{varma2023explaining} argue the converse: grokking occurs when the optimizer converges to an ``efficient" circuit which generalizes well, from which point weight decay can reduce the parameter norm without harming accuracy. This work bridges the gap towards a causal understanding of the relationship by shedding light on the mechanisms by which the parameter norm can influence the onset of grokking in Section~\ref{sec:grokking}, finding some truth in both perspectives.

\subsection{Optimization Dynamics and the Effective Learning Rate}
A crucial feature of our analysis in the sections that follow concerns the control of the \textit{effective learning rate}, a quantity which has significant implications on learning dynamics~\citep{arora2018theoretical, li2020exponential} in scale-invariant functions. While the learning rate has a large effect on how gradient-based optimizers perform in practice, this effect depends on the norm of the parameters and the gradients. The aim of the effective learning rate is to provide a metric independent of these quantities.
Concretely, let $f$ be a scale-invariant function for some $\alpha \neq 0$; that is, assume that  $f(\theta) = f(\alpha \theta)$ for all $\theta$ in its domain. Then,
\begin{equation}
    f(\theta + \eta \nabla f(\theta)) = f(\alpha \theta + \alpha^2 \eta \nabla f(\alpha \theta))  \;,
\end{equation}
and hence the effective learning rate $\tilde{\eta}$ to replicate the learning dynamics on $\theta$ with unit-norm parameters $\tilde{\theta}$ is given by $\tilde{\eta}(\theta) = \frac{\eta}{\|\theta\|^2}$ for gradient descent {(see e.g. \citet[Definition 1]{lyle2024normalization} for a derivation of this relationship)} and $\tilde{\eta}(\theta) = \frac{\eta}{\|\theta\|}$ in adaptive optimizers like RMSProp and Adam which approximate the update $\frac{\nabla f(\theta)}{\|\nabla f(\theta)\|}$.
Such a rescaling is exact for scale-invariant functions, but even in networks which are only partially or approximately scale-invariant, such as those which include normalization layers, increased parameter norm is empirically associated with reduced sensitivity to gradient updates and loss of plasticity~\citep{lyle2024disentangling, nikishin2022primacy, dohare2023loss}. The relationship between parameter norm and effective learning rate is leveraged by \citet{lyle2024normalization} in the Normalize and Project (NaP) algorithm, a method which aims to improve robustness to nonstationarity by maintaining tight control on the effective learning rate. NaP inserts normalization layers prior to each nonlinearity in the network and periodically projects the weights in each linear layer to the unit sphere (w.r.t. Frobenius norm $\| \cdot \|_F$), resulting in the two-step update for the  matrix-valued parameters $W^\ell$ of each layer $\ell$, where $u^\ell_t$ denotes the optimizer update to $W^\ell$:
\begin{equation}\label{eq:renormalization}
   \tilde{W}^\ell_t \gets W^{\ell}_{t-1} + \eta_t u^\ell_t, \quad  W^\ell_t \gets \tilde{W}^\ell_t \frac{\|W^\ell_0\|_F}{\|\tilde{W}^\ell_t\|}~.
\end{equation}
In some cases, it can be desirable to apply the projection step only every $k$ updates either to reduce the computational overhead of the method or as an analytical means of interpolating between projection and regularization of the parameter norm. We will specify in later sections when this is the case, and default to projection every step.
We will use this approach to ensure that the effective learning rate can be directly read off from the explicit learning rate schedule in later sections.

\section{Feature-learning dynamics}
\label{sec:feature-learning}

{This section introduces the empirical and methodological tools which we will use to study the emergence of feature-learning in neural network training. Section~\ref{sec:fl-metrics} will present a set of empirical metrics which we will use to determine whether a network is meaningfully changing its learned representation, and Section~\ref{sec:fl-elr} will propose a simple learning rate re-warming strategy to induce feature-learning dynamics on demand.}

\subsection{Feature-learning metrics}
\label{sec:fl-metrics}
The dichotomy between kernel and feature-learning dynamics was first delineated in the study of infinite-width neural networks \citep{xiao2020disentangling, yang2019wide}, where for example \citet{yang2022tensor} define feature-learning as an $\Omega(1)$ change in the limiting feature covariance matrix of an intermediate layer of the network from its initial value as the width increases. Under this definition, any finite-width network vacuously perform some amount of feature learning. What matters in practice, however, is the \textit{magnitude} of the change undergone by the features, a quantity whose characterization admits a number of reasonable metrics. One such metric, inspired by the definition of~\citet{yang2022tensor}, is the change in the normalized feature covariance matrix at a given layer after some interval $T$. In particular, letting $f^\ell$ denote the output of layer $\ell$ and $\mathbf{X}$ denote some set of $n$ training datapoints, with $f^\ell(\mathbf{X}) \in \mathbb{R}^{n \times d}$ denoting the matrix obtained by stacking the outputs for each datapoint, we define
\begin{equation}\label{eq:delta_C}
 \Delta_C^\ell(t, T) \overset{\mathrm{def}}{=} \|C^\ell_t(\mathbf{X})  - C^\ell_{t+ T}(\mathbf{X}) \| \quad \text{ where } C^\ell_t(\mathbf{X}) =  \frac{f_t^\ell(\mathbf{X}) f_t^\ell(\mathbf{X})^\top}{\|f_t^\ell(\mathbf{X})\|^2} \; .
\end{equation}
While exhibiting a number of appealing properties such as rotation-invariance and grounding in the literature, this metric does not capture the degree to which the network is taking advantage of the nonlinearities in the activation functions. As the networks we consider in this work primarily use ReLU nonlinearities, we use changes in the \textit{activation pattern} $A$~\citep[a matrix of binary indicators of non-zero activations,][]{poole2016exponential} produced by a hidden layer in the network as an additional proxy for feature-learning, defining
\begin{equation}\label{eq:delta_A}
    \Delta_A^\ell(t, T) \overset{\mathrm{def}}{=} \|A^\ell_t(\mathbf{X})  - A^\ell_{t+ T}(\mathbf{X}) \| \quad \text{ where } A^\ell_t(\mathbf{X})[i, j] = \delta(f^\ell(\mathbf{X}_i)_j > 0)\; ,
\end{equation}
where $\delta(E)$ denotes the indicator function of an event $E$.
We will primarily use these two metrics in the sections that follow to quantify the rate of feature-learning in a network.

\subsection{Re-warming the effective learning rate}
\label{sec:fl-elr}
Theoretical analyses of neural networks in the infinite-width limit characterize the dynamics of a neural network at initialization in terms of three quantities: the initial parameter norm, the learning rate, and the scaling factor by which a layer's outputs are multiplied~\citep{everett2024scaling}. For example, the neural tangent kernel regime~\citep{jacot2018neural} maintains a fixed kernel throughout training, enabling $O(1)$ change in the network outputs by balancing the vanishing parameter updates with growth in the network Jacobian, while other parameterizations~\citep{yang2022tensor, sohl2020infinite} allow not only the final network outputs but also the learned representations to evolve. The ratio between the learning rate and parameter norm thus plays an important role in determining the regime of the training dynamics at initialization. In non-stationary learning problems, however, it is not sufficient to only consider the initialization; one must also ensure that suitable dynamics can be recovered as needed throughout training in response to changes in the learning problem. 

\begin{algorithm}[H]
\caption{{Adaptive ELR Re-Warming}} \label{alg:cap}
\begin{algorithmic}
\State \textbf{Input:} scale-invariant network $f$, initial parameters $\theta_0$, budget $T$, loss $\ell$, data distribution(s) $(\mathcal{P}_t)_{t=1}^T$, optimizer \texttt{Update}, optimizer state $S_0$, 
\For{$1 \le t \leq T$}
\State sample $\bX_t \sim \mathcal{P}_t$
\State $\theta_{t+1}, S_{t+1} \gets$ \texttt{Update}($\theta_{t}, \bX_t, S_t)$
\State $\theta_{t+1} \gets \texttt{Project}(\theta_{t+1})$
\If{\texttt{rewarm}$(\theta_t, \bX_t, S_t)$  \; \# e.g. \texttt{rewarm}$(\theta_t, \bX_t, S_t)$ =  CUSUM($\ell(\theta_t, \bX_t$))} 
\State $S_{t+1}$ $\gets$ \texttt{ResetLR}($S_{t+1}$) 
\EndIf
\EndFor
\end{algorithmic}
\end{algorithm}

To achieve this adaptivity, we propose a simple modification to the Normalize-and-Project (NaP) method of \citet{lyle2024normalization}. Rather than simply avoiding decay in the ELR, as was the method's original intent, we propose to use NaP to periodically \textit{increase} the ELR to facilitate more rapid feature-learning, a technique we will refer to throughout this paper as \textbf{ELR re-warming}. This approach can be easily adapted to any gradient-based optimization algorithm: after each optimizer update step, we re-scale the parameters to their initial norm and update the optimizer's learning rate based on either a fixed schedule (for example a cyclic schedule with a pre-specified frequency, \citealp{smith2017cyclical}) or an adaptive criterion (e.g., CUSUM of \citealp{page1954continuous} applied to the loss). We provide pseudocode in Algorithm~\ref{alg:cap}, and give a more rigorous discussion of the role of the learning rate on feature-learning dynamics in Appendix~\ref{appx:theory}.

\section{Grokking in modular arithmetic tasks}
\label{sec:grokking}
The onset of grokking in modular arithmetic tasks presents a microcosm of feature learning in neural networks. 
While prior works have extensively studied qualitative properties of the representation before and after grokking \citep{liu2022towards, varma2023explaining}, we focus on \textit{using} grokking as an indicator of feature-learning in the network. We first show that the effective learning rate plays a critical role in the emergence of grokking provided that a suitable network parameterization is used. We then demonstrate that our ELR re-warming approach can induce grokking on-demand at any point in training, and that learning rates which induce grokking are also those which induce more rapid changes in the learned features.

\textbf{Experiment setting.}
We train a transformer with a single attention block on a modular arithmetic task, replicating the architecture of \citet{varma2023explaining} (we provide additional details on the model architecture in Appendix~\ref{sec:exp-details-grokking}). The task in question consists of a set of inputs of the form $x \cdot y = \square$, where $\square$ is a blank token which must be predicted by the model, and where the correct answer is of the form $x + y \mod 117$. To generate a train-test split, we randomly extract a fraction $\rho=0.2$ of the set of all $x, y$ pairs, with $0 \leq x, y < 117$; the network is trained on this dataset for 1M steps, and evaluated on the remaining set of $(x, y)$ pairs not used in training. We do not plot the training accuracy of these networks due to space constraints, but note that almost all networks attain perfect or near-perfect training accuracy almost immediately after training begins (typically within one thousand steps). We note explicitly instances where this is not the case.
\begin{figure}
    \centering
  \includegraphics[width=0.5\linewidth]{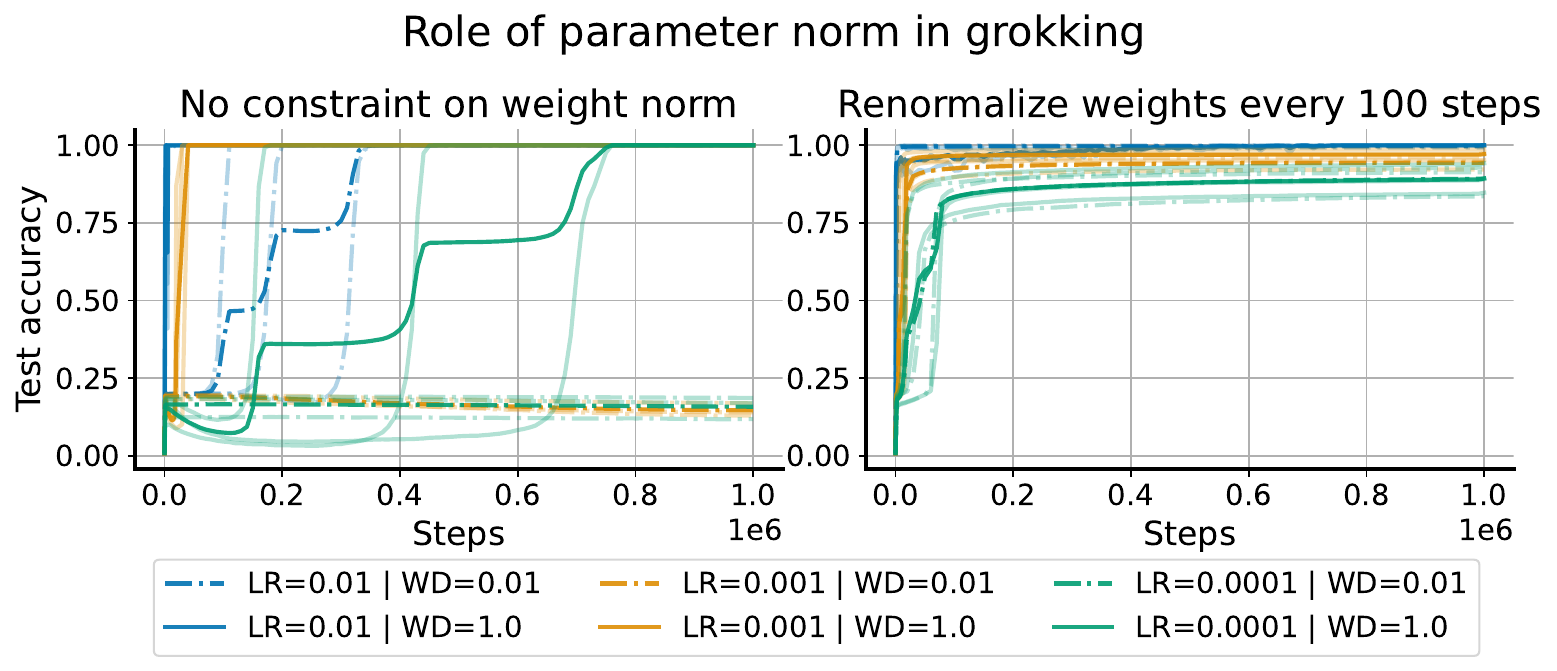}
    \includegraphics[width=0.49\linewidth]{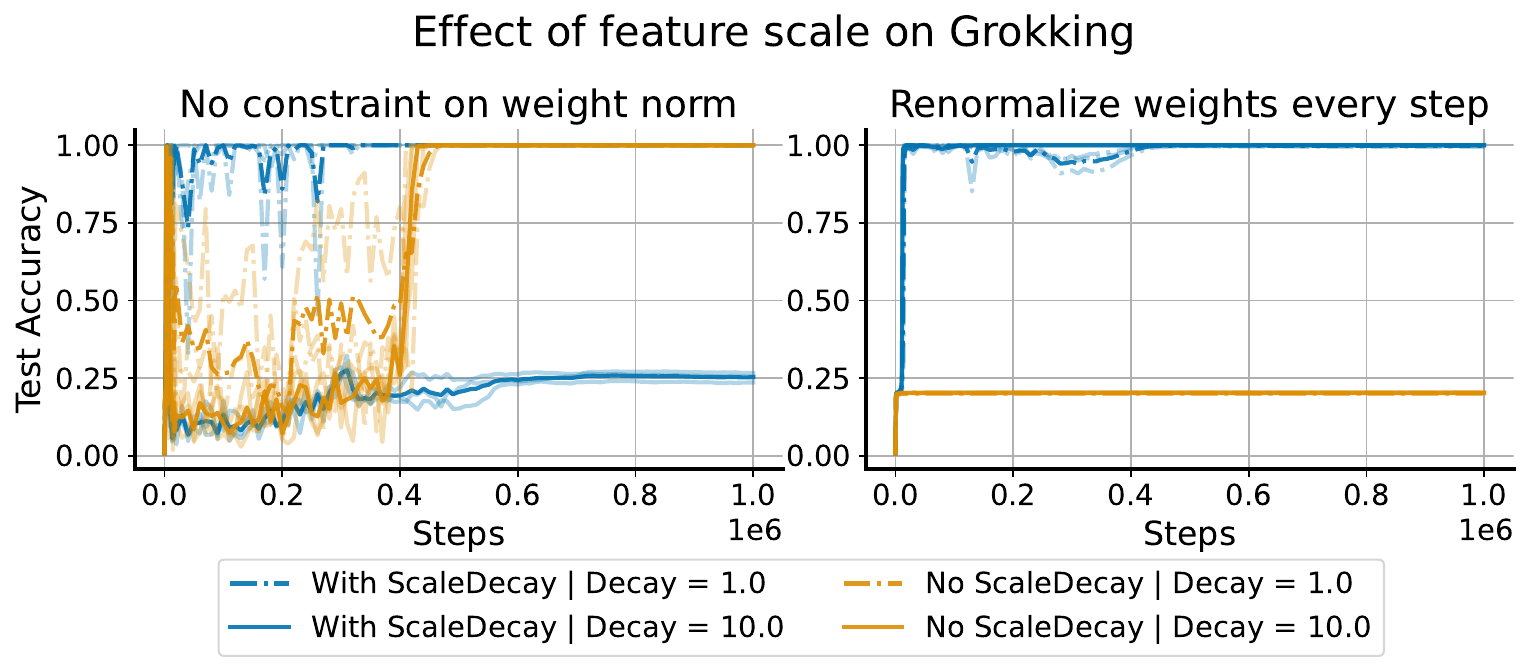}
    \vspace{-0.3cm} 
    \caption{\textbf{LHS:} grokking occurs in networks whose updates induce nontrivial changes in the learned features, a property which requires a sufficiently large update norm relative to the parameters. This can be achieved by increasing the learning rate or decreasing the parameter norm via weight decay, where a larger weight decay can compensate for a smaller learning rate and vice versa. Periodically re-scaling the parameters to their original norm thus blunts the effect of weight decay. \textbf{RHS:} when we add layer normalization to the network, a large ELR is not sufficient to induce grokking. Instead, it becomes necessary to also reduce the norm of the attention head inputs, which we achieve by applying weight decay to the scale parameters of the layernorm transforms in the network.}
    \label{fig:wd-lr-figure}
\end{figure}

\subsection{Training dynamics of delayed generalization}

A variety of prior works have emphasized the critical role of weight decay in grokking. However, 
it is not clear what direction the causal arrow points -- was it the smaller parameter norm that facilitated a better-generalizing solution, or did the network's discovery of a generalizing solution allow it to reduce the parameter norm without harming accuracy?
{Any attempt to answer this question will be further confounded by the relationship between the parameter norm and effective learning rate. Indeed, based on our previous arguments, we predict that a large effective learning rate will induce feature-learning dynamics and thus quickly facilitate grokking even if the parameter norm does not decrease below its initial value.
We design an experiment to isolate the role of the effective learning rate in grokking by sweeping over learning rate and weight decay values in the experimental setting described above.} We perform this sweep in two settings: one where the parameter norm is allowed to vary, and one where parameters are periodically rescaled to maintain a roughly constant norm throughout training. We use the same network architecture, which does not incorporate layer normalization, in both cases. We observe in Figure~\ref{fig:wd-lr-figure} (left hand side) that periodic weight projection results in rapid and consistent generalization across random seeds, a feat which networks with variable parameter norm require a large weight decay parameter to achieve. Further, we see a stronger effect of learning rate compared to weight decay value on grokking in the variable-norm networks, again reinforcing that the effective learning rate, and not the parameter norm on its own, is responsible for grokking.

While a high effective learning rate is one mechanism by which networks achieve feature learning, inducing feature-learning dynamics alone, for example by moving from one poorly generalizing solution to another, is not sufficient to achieve good test set performance. In the right subplot of Figure~\ref{fig:wd-lr-figure} we provide an example of this situation by adding layer normalization to the network architecture. Despite following an otherwise identical experimental protocol, we obtain wildly different results: in this new architecture, weight normalization completely fails to generalize. The reason for this is subtle: the layer normalization transform rescales the attention inputs to be much larger than they would be under the default weight initialization, and as a result the attention matrices have much lower entropy and are more difficult to optimize~\citep{wortsman2023small}. If the network is able to modulate the parameter norm, it can increase the smoothness of the attention mask by reducing the norm of the key and query matrix parameters; the projected network has no such means of escape. However, if we apply a weight decay term to the scale parameters of the normalization layers,a technique which we refer to as which we refer to by the term `scale decay', we obtain the blue curves on the right hand side of Figure~\ref{fig:wd-lr-figure}. We thus conclude that generalization requires \textit{both} meaningful changes to the learned representation, and a suitable loss landscape in which these changes converge to a smooth, generalizing solution.

\subsection{ELR re-warming accelerates generalization}
\begin{figure}
  \includegraphics[width=\linewidth]{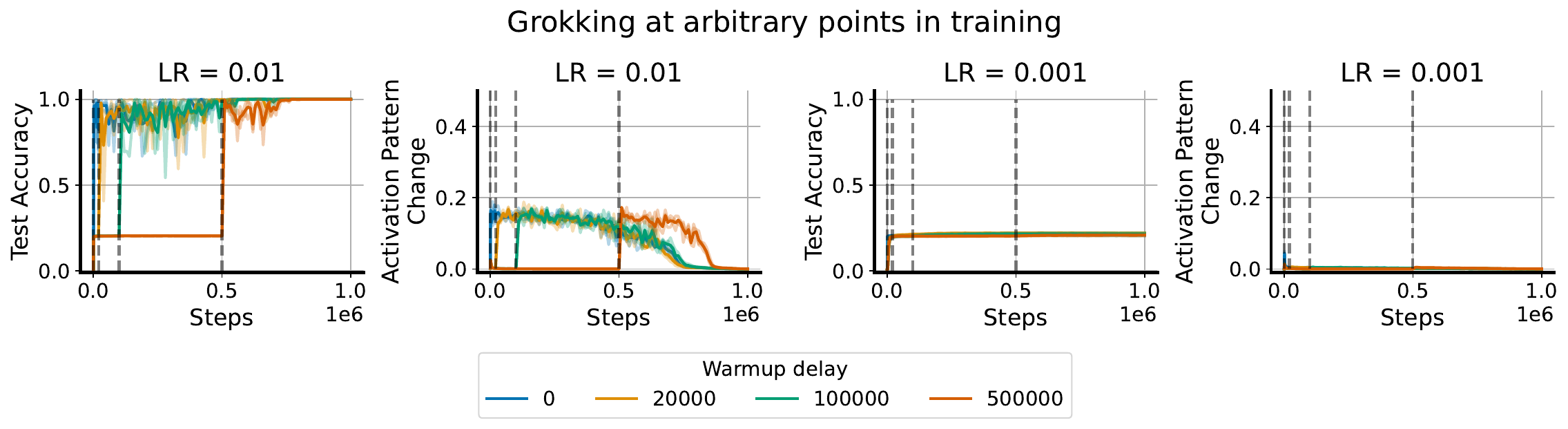}
     \vspace{-0.6cm} 
\caption{We can induce grokking at arbitrary points during training using targeted increases in the effective learning rate (timesteps where a learning rate increase occurs are denoted by dotted lines). This approach requires that the learning rate increase be sufficient to move feature learning metrics such as the percentage change in activation patterns in the network -- we see in the LHS that increasing the learning rate to 0.01 leads to feature learning and grokking, whereas on the RHS 0.001 does not.}
\label{fig:delayed-grokking}
\end{figure}

Having identified the importance of the effective learning rate in grokking, we now explore how it can be manipulated to accelerate the onset of generalization. In order to achieve generalization, the optimization process much satisfy two criteria. First, the optimizer must spend sufficient time in a regime where nontrivial updates are being made to the network.
Second, once generalization is achieved the learning rate should be reduced to improve stability and In Figure~\ref{fig:delayed-grokking} we apply the protocol which generated the blue line of the rightmost subplot of Figure~\ref{fig:wd-lr-figure}: we train a transform with weight projection, layer normalization, and scale decay. In particular, by increasing the effective learning rate to a sufficiently large value via linear warmup and then following a cosine decay schedule, we rapidly obtain perfect test accuracy which persists even after one million optimizer steps. 

An important difference between the typical setting for grokking and the continual learning settings which we will next investigate is that the modular arithmetic problem is stationary, and while we have shown how to accelerate grokking, have not shown that it is possible to induce grokking at arbitrary points in the learning process. It is plausible that the process by which features are learned depends crucially in some way on changes which our approach makes during the earliest phase of training. Figure~\ref{fig:delayed-grokking} addresses this concern. We consider a range of increasingly long offsets, during which training at a learning rate too low to induce grokking occurs. After the desired interval, we then `turn on' the learning rate schedule and proceed with training as normal. We observe that even after several hundred thousand optimizer steps, it is still possible to induce grokking via ELR re-warming.

\section{Warm-starting neural network training}
\label{sec:warm-starting}
We now explore whether ELR re-warming can also facilitate feature-learning, and hence generalization, in a different neural network architecture, data modality, and training regime: warm-starting image classifier training. Whereas a random initialization is designed to be relatively easy to overwrite, in this new setting the network must learn to overwrite previously learned features, a task which has the potential to be significantly more difficult than the grokking regime studied previously.

\textbf{Experiment details.}
In this section, all experiments are run with convolutional architectures on the CIFAR-10 dataset. Denoting this dataset $\mathcal{D}_{\mathrm{total}}$, we expand on the protocol of \citet{ash2020warm} by first training the network on some subset $\mathcal{D}_{\mathrm{init}}$ which contains a randomly sampled subset of the data points in $\mathcal{D}_{\mathrm{total}}$. After a fixed interval of 70 epochs, the remainder of the dataset is added to $\mathcal{D}_{\mathrm{init}}$, and the network continues training on the full dataset for another 70 epochs. We track test accuracy throughout the entire procedure. As a baseline, we include one setting where we ``pretrain" on 100\% of the training data, which presents an upper bound on the expected performance improvement from our approach.
\begin{figure}
    \centering
    \includegraphics[width=\linewidth]{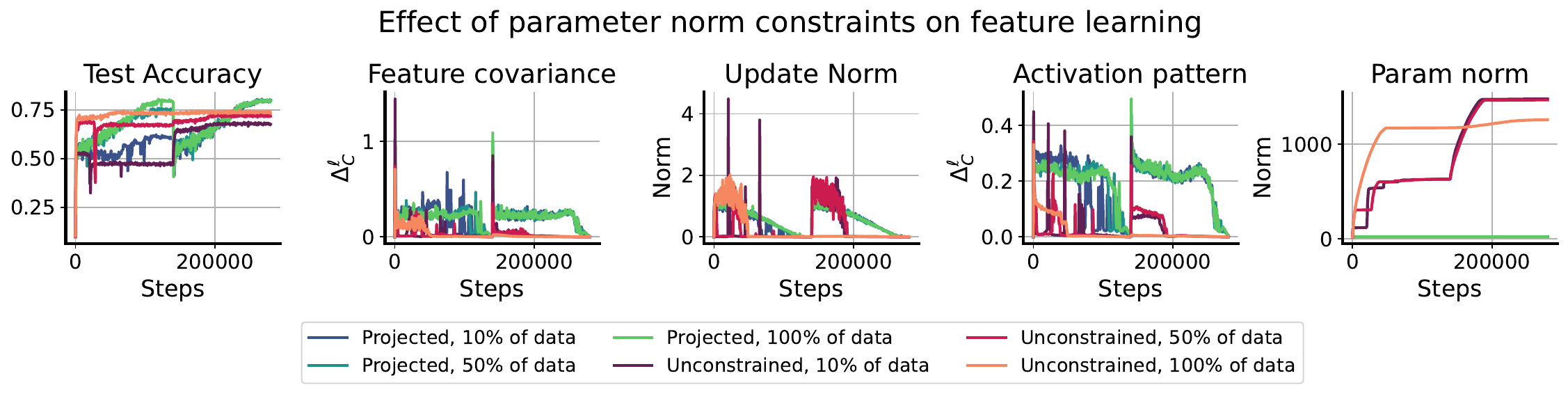}
        \vspace{-0.6cm} 
    \caption{Learning rate re-warming on networks whose parameter norm is constrained so as to not grow significantly over the course of training exhibit significant improvements on CIFAR-10 over naively applying learning rate cycling on an unregularized network. We observe that feature-learning measures increase and then decline with each learning rate cycle in tandem with the learning rate. Although it applies similar update norms to the parameters, the unconstrained network exhibits notably less feature-learning in the second iteration by both the feature covariance and activation pattern metrics.}
    \label{fig:cyclic-warmstarting}
\end{figure}

\subsection{ELR re-warming closes the generalization gap}

{The Shrink and Perturb method~\citep{ash2020warm} is a popular means of improving generalization on non-stationary tasks~\citep{schwarzer2023bigger}. This approach, which involves rescaling the parameters by a constant $0 < \alpha < 1$ and then sampling a perturbation from the initialization distribution, can be viewed as a means of reducing the difficulty of escaping the local basin of the loss landscape at the current learning rate.}
Similar observations have been made theoretically in non-stationary online convex optimization, where the ``effective diameter'' of the active parameter space determines how quickly one can adapt to new data distributions and, among others, similar techiques are used to control the diameter \citep[see, e.g.,][]{GySz16}. We propose to instead escape from a local minimum not by changing the loss landscape, but by increasing the learning rate, for example to a value which exceeds the sharpness of the local basin~\citep{lewkowycz2020large}, so that the optimizer has the chance to jump out of the current basin.
In Figure~\ref{fig:cyclic-warmstarting}, we test whether ELR re-warming can close the generalization gap that emerges in warm-started neural networks, following the procedure outlined above (additional details can be found in Appendix~\ref{sec:exp-details-warm}). We use five random seeds on three initial dataset fractions, corresponding to 10\%, 50\%, and 100\% of the whole dataset.
We track both notions of feature learning described in Equations~\ref{eq:delta_C} and~\ref{eq:delta_A}, and observe that while the update norm for both training setups is similar in both iterations of training, the NaP variant applies more significant changes to the learned representation particularly in the second iteration, and attains the same final generalization performance independent of the initial dataset size. 

In the rightmost plot of Figure~\ref{fig:per-layer-lrs} we ablate the two critical components of the adaptive re-warming process: the use of weight projection to prevent unwanted decay in the effective learning rate over time, which would reduce the ability of the network to change its learned representation, and the use of a changepoint detection algorithm (for example the Cusum algorithm, \citealp{page1954continuous}) on the loss value to identify non-stationarities in the data-generating distribution which merit a learning rate reset. We show that removing any single component of this approach degrades performance, with the unconstrained parameters trained on a single continuous learning rate decay performing worst. We further show in Appendix~\ref{sec:appx-cyclic-warmstarting} that these results are also achieved when the fixed schedule does not align with the task change.

\subsection{Learning dynamics analysis and per-layer learning rates}

\begin{figure}
    \centering
      \includegraphics[height=3.7cm]{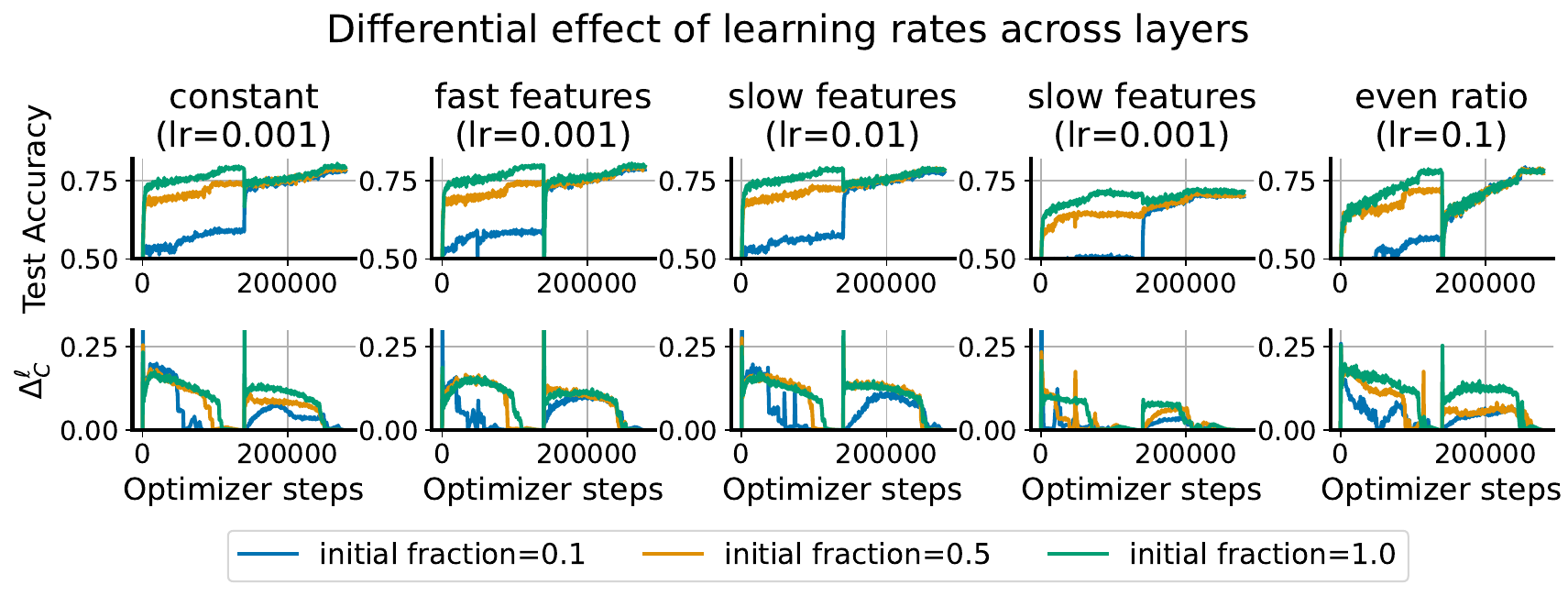}
      \includegraphics[height=3.7cm]{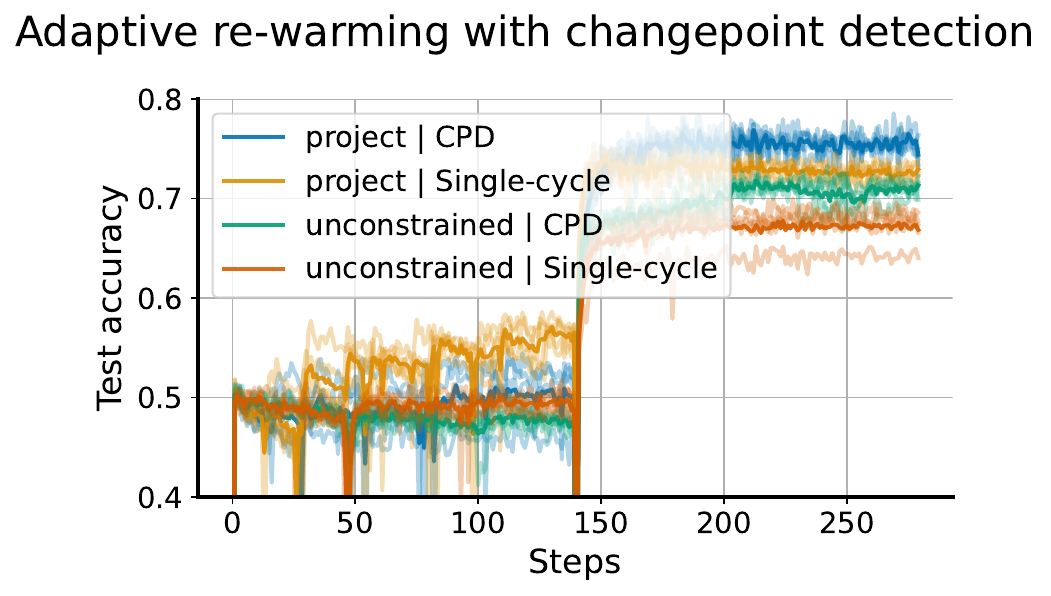}
    \vspace{-0.3cm} 
    \caption{\textbf{LHS: per-layer learning rate scaling.}  Re-scaling learning rates in different layers results in significant changes in the magnitude of the warm-starting effect. Intriguingly, while conditions with a higher absolute number of dead neurons exhibit larger warm-starting effects, observing a higher relative number of dead units in networks warm-started from smaller datasets does not indicate that there will be a large generalization gap. \textbf{RHS:} using the CUSUM changepoint detection algorithm to determine when to re-warm the learning rate successfully closes the generalization gap in the warm-starting experiment regime; however, to gain the full benefits of learning rate rewarming, the parameter norm must be prevented from growing during the initial learning phase.} 
    \label{fig:per-layer-lrs}
\end{figure}

Whereas the transformer architectures we considered in Section~\ref{sec:grokking} consisted of a single attention layer, the deeper vision architectures allow us to investigate whether different layers might exhibit different dynamics which benefit from distinct learning rates.
Prior work suggests that different layers in vision networks can exhibit distinct dynamics~\citep{zhang2019all} and train optimally at distinct learning rates~\citep{everett2024scaling}. While these findings are most applicable to networks far larger than the ones we consider in this work, we conduct a preliminary investigation to gauge the utility of layer-specific learning rates in maintaining plasticity, whose results we present in the LHS of Figure~\ref{fig:per-layer-lrs}.
We consider four different per-layer learning rate assignments: a \textit{constant} assignment, where every layer is given the same learning rate; \textit{fast features}, where the final linear layer's learning rate is divided by 10; the analogous \textit{slow features}, where the non-final layers' learning rates are divided by 10; and \textit{even-ratio-abs}, which scales the learning rate of each layer proportional to the average magnitude of the parameters, so that updates applied to each layer correspond to a constant proportion of the layer's L1 norm. The choice to distinguish between the final layer and all prior layers in two of these assignments is due to the previously observed importance of the final layer scale in facilitating grokking in some vision tasks~\citep{liu2022omnigrok}. We find that provided the learning rate on the \textit{features} component of the network is held constant, the learning rate on the final linear layer does not have a noticeable effect on performance at the magnitudes we evaluated.

\section{Reinforcement Learning}
\label{sec:reinforcement-learning}
Reinforcement learning presents a particularly challenging non-stationary learning problem due to the pathological dynamics induced by bootstrapping~\citep{van2018deep}, the nonstationarity in the input observations as the state-visitation distribution evolves, and nonstationarity in the prediction target as a result of policy improvement~\citep{lyle2022learning}. While this nonstationarity makes RL a promising candidate to benefit from methods which accelerate feature-learning, its notoriety for instability and divergence~\citep{ghosh2020representations} means that any attempt to increase the volatility of the learning dynamics, as we have sought in our ELR re-warming strategies, must be undertaken with care. In particular, we must ensure that by accelerating the rate at which the network changes its learned representation, we do not also accelerate its divergence or collapse before a good policy can be learned. We conjecture that learning rate \textit{cycling} might therefore be particularly well-suited to reinforcement learning. By periodically increasing (re-warming) the learning rate, we aim to overwrite spurious correlations which interfere with policy improvement, and by then annealing the learning rate to a stable value we can avoid the instabilities that so often plague RL algorithms. We include additional details on the experiments conducted in this section in Appendix~\ref{sec:exp-details-rl}.

\subsection{Cyclic learning rates in online reinforcement learning}
\begin{figure}
    \centering
    \begin{minipage}{0.425\textwidth}
\centering
\includegraphics[width=0.9\linewidth]{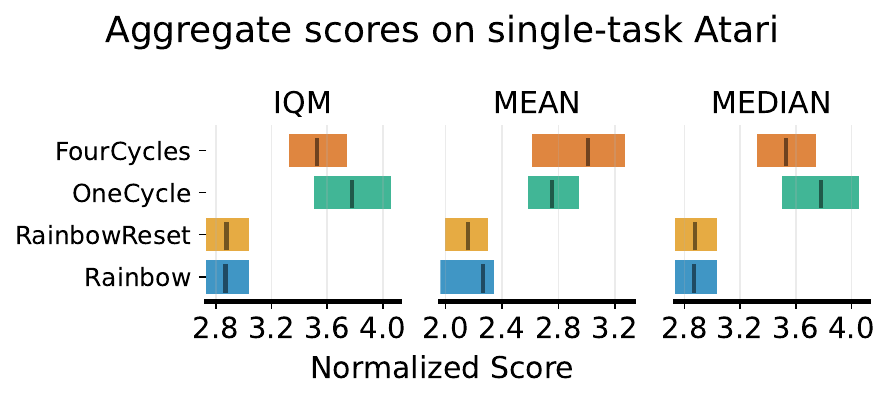}
\includegraphics[width=1.0\linewidth]{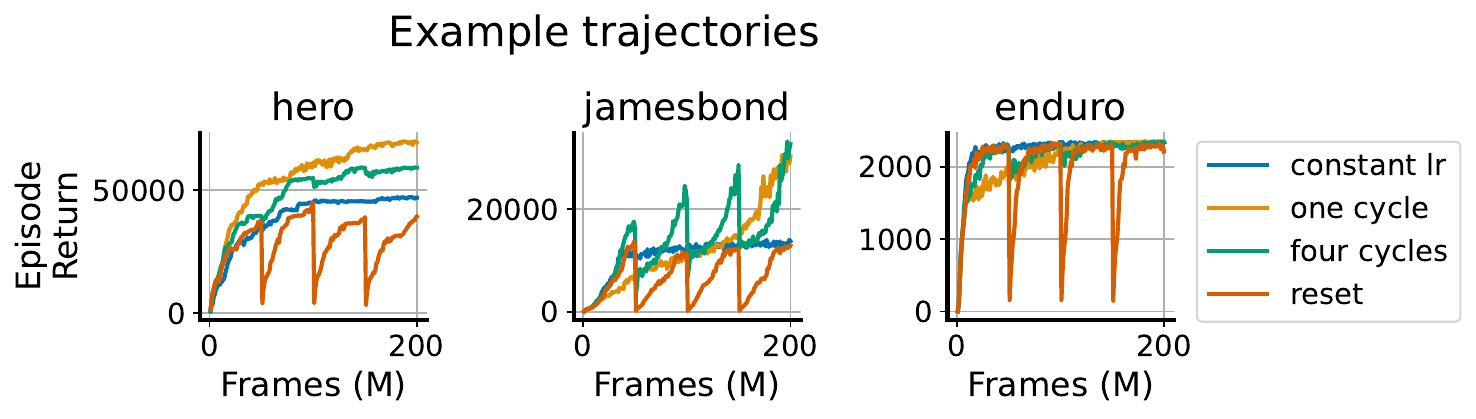}
\end{minipage}
\begin{minipage}{0.57\textwidth}
    \includegraphics[width=1.0\textwidth]{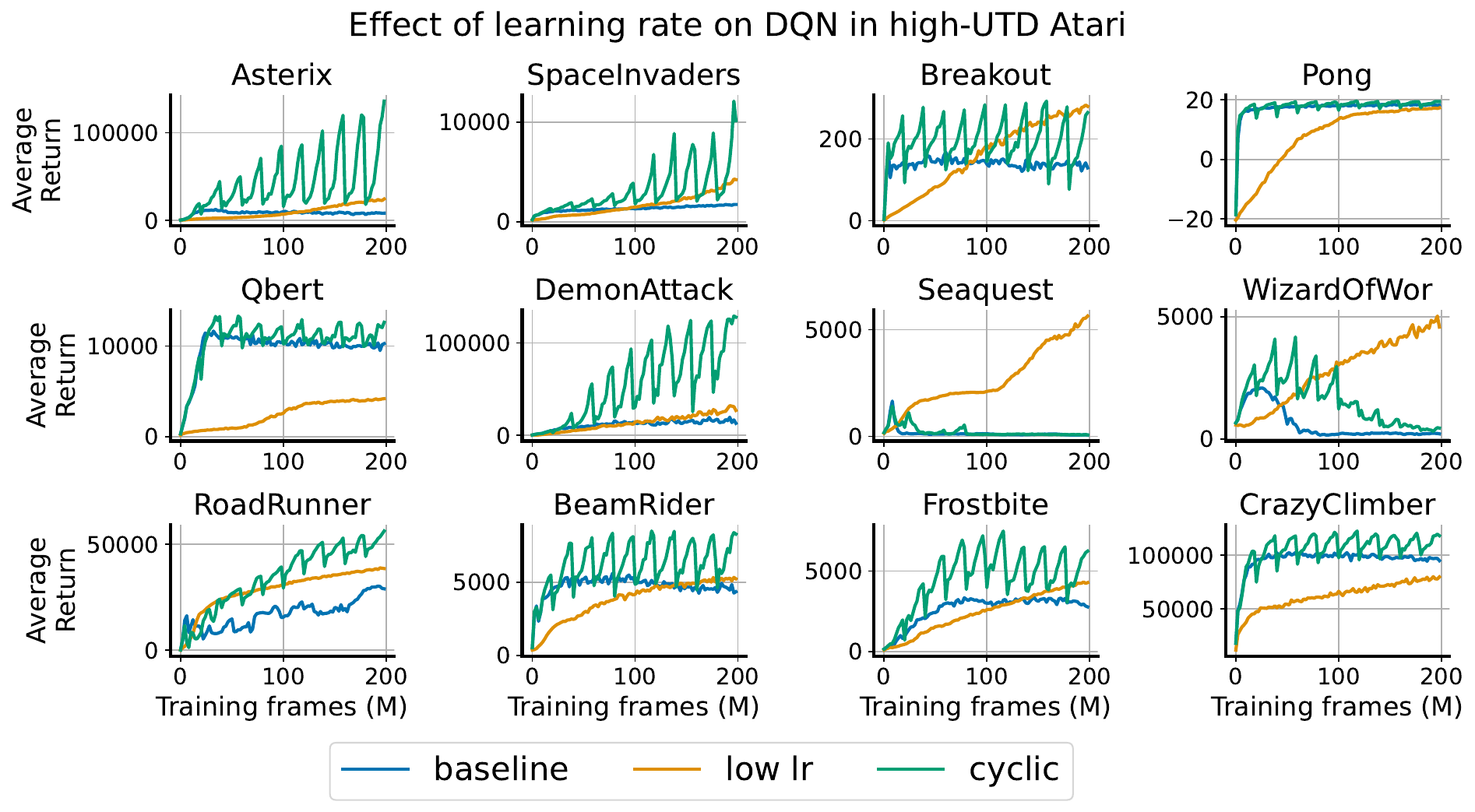}
\end{minipage}

    \vspace{-0.3cm} 
\caption{\textbf{LHS:} Cycling the learning rate in a Rainbow agent with weight projection trained in a low update-to-data regime exhibits heterogeneous effects across environments, outperforming resets and a constant learning rate but exhibiting similar performance to a single cycle of learning rate decay. Agents recover faster from learning rate re-warming than from resets. \textbf{RHS:} Cyclic learning rates produce much more pronounced effects in DQN under a high replay ratio, outperforming both fixed learning rate baselines in several environments, but sometimes suffers from instability in environments like Seaquest and Wizard of Wor.}
    \label{fig:cyclic_atari}
\end{figure}

We begin our investigation by considering cyclic learning rates in two different training regimes: a high update-to-data (UTD) ratio or \textit{data-efficient} regime where the environment and learner take steps in tandem, and the low update-to-data ratio or \textit{online} regime, where the environment takes four steps per every learner update step. The data-efficient regime, which often uses much higher ratios of learner updates to environment steps than that considered in this paper, is notoriously prone to over-estimation bias and feature collapse, particularly when the training algorithm uses bootstrapping, and benefits significantly from a variety of resetting strategies. By contrast, the online regime tends to be more stable and benefits less if at all from parameter resets. Because learning rate re-warming bears many similarities to resets, we conjecture that it should provide a greater benefit in the high UTD regime. We test these hypotheses with a DQN \citep{mnih2015human} and a Rainbow agent \citep{hessel2018rainbow} in the arcade learning environments (ALE) \citep{bellemare2013arcade} with our results presented in Figure~\ref{fig:cyclic_atari}.

\paragraph{Low UTD regime} We observe heterogeneous effects from learning rate cycling across environments in the LHS subplot of Figure~\ref{fig:cyclic_atari}, where we present results for Rainbow agents trained with NaP. While the resulting performance curves from cyclic learning rates exhibit some similarity to those attained by resets in that both exhibit an initial performance drop, the cyclic LR agent recovers more rapidly and attains higher final performance on most environments. We tend to see convergence of the cyclic schedules to approximately the same performance at the end of training regardless of the number of cycles applied, suggesting that the benefits of these approaches can be attributed largely to spending sufficient time at key learning rates rather than exposing the network to higher ELRs later in training. 

\paragraph{High UTD regime} Whereas in the low UTD regime performance of all methods frequently converged to the same final value, we see much greater variability in score in the high-UTD regime, largely due to the instability of larger learning rates. We evaluate two different learning rates on a DQN agent trained without weight decay or projection, where the \textit{baseline} value is trained at a constant learning rate of 6.25e-5, while the \textit{low lr} variant trains with a constant step size of 1e-6, along with a cyclic variant that oscillates between the two values. While the cyclic variant often exhibits performance improvements, it suffers from similar degrees of instability in environments such as \textit{seaquest} and \textit{krull} as the higher learning rate variant.
In the next section, we will investigate a means of mitigating this instability.

\subsection{Combining cyclic learning rates with annealing}
\begin{figure}
    \centering
    \includegraphics[width=0.49\linewidth]{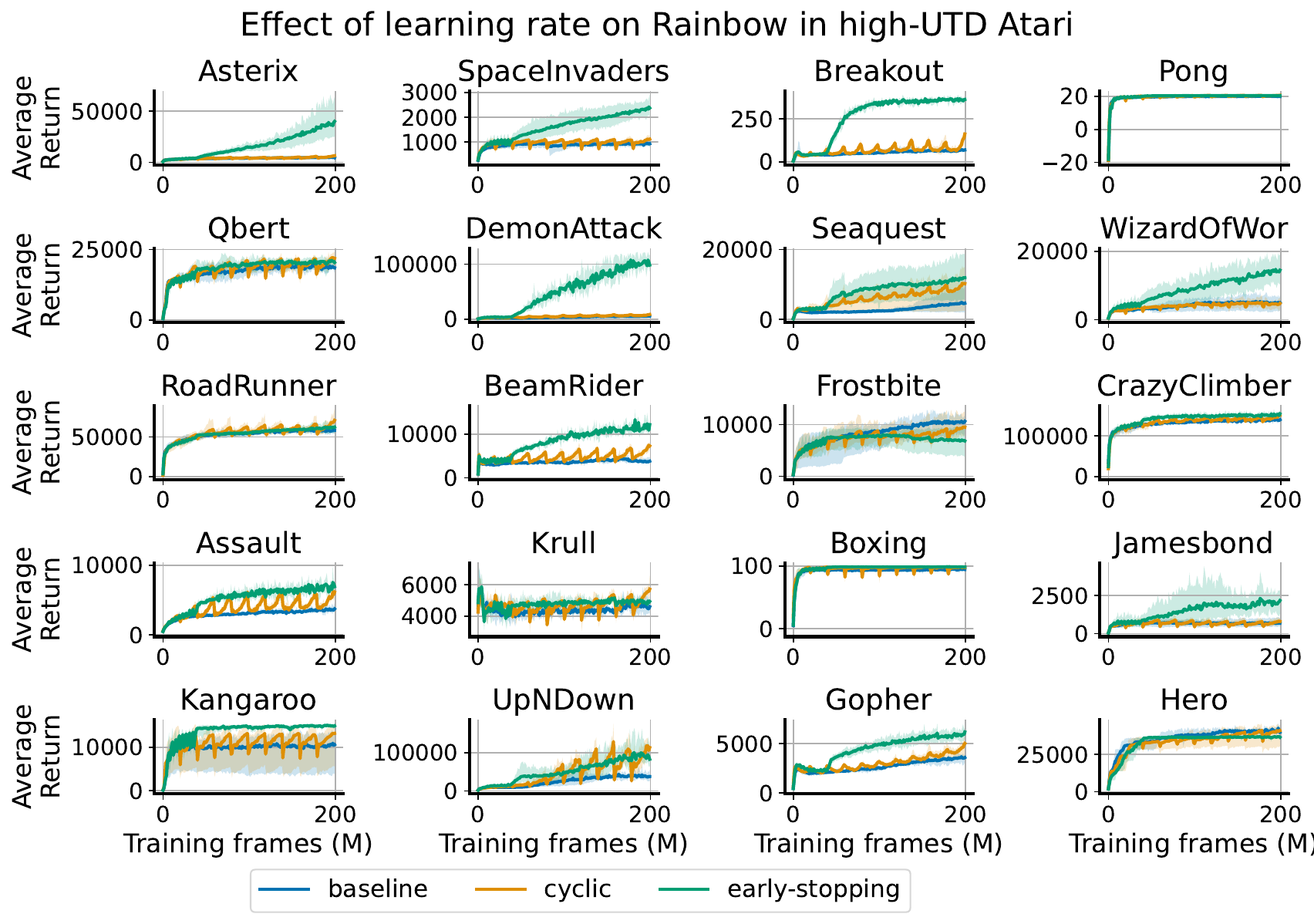}
    \includegraphics[width=0.49\linewidth]{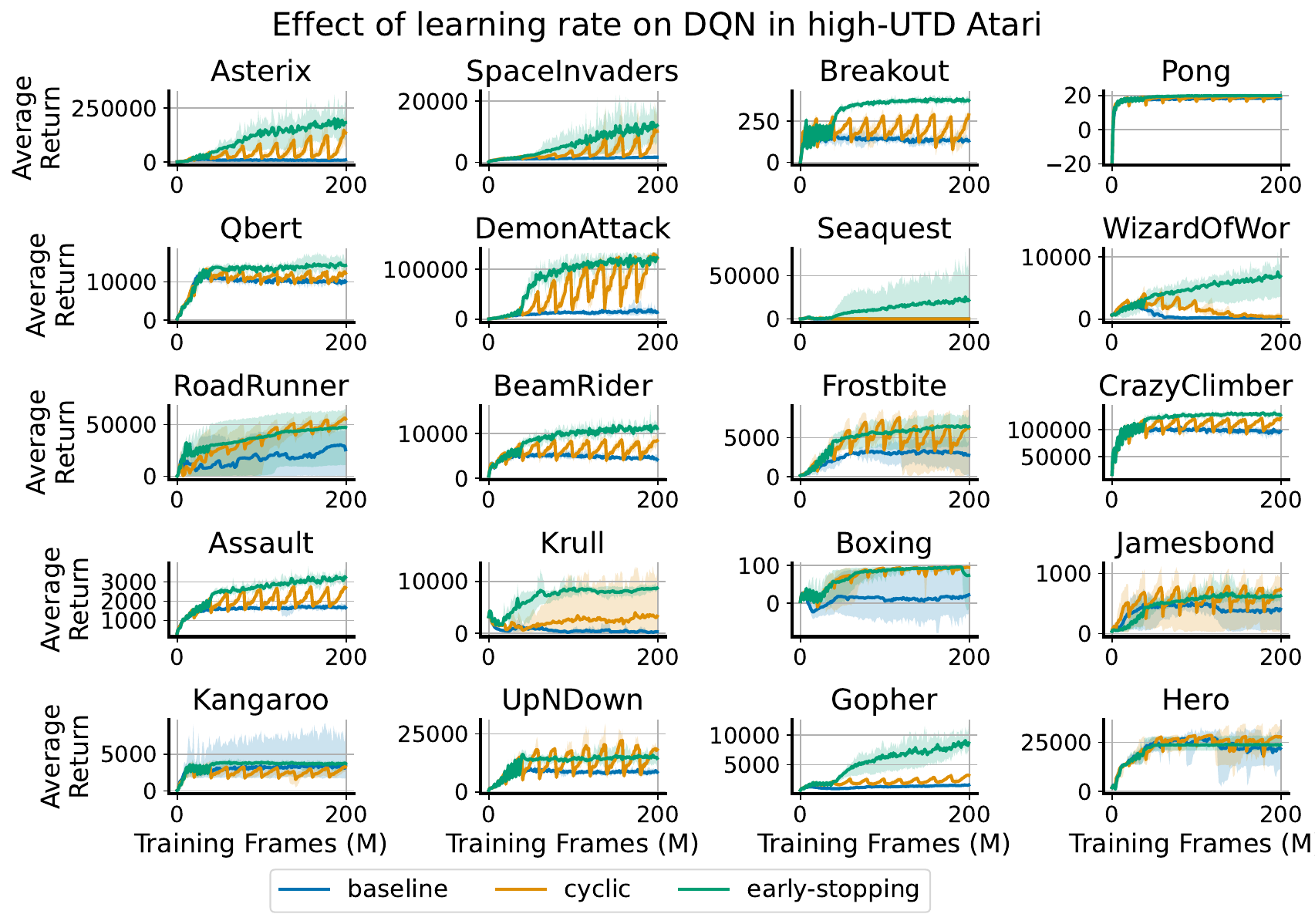}
        \vspace{-0.3cm} 
    \caption{DQN and Rainbow agents both benefit from early learning rate cycling followed by annealing in high-UTD Atari settings, with significant improvements over the constant and continual cyclic learning rates.}
    \label{fig:es-cycling}
\end{figure}

While LR cycling improves performance in some environments in the high-UTD regime, it is clear that in many environments optimal performance requires extended periods of low-learning-rate training. These difficulties can in large part be attributed to the nature of the environments on which we evaluate: reinforcement learning agents tend to experience greater non-stationarity at the beginning of training, and also tend to see significant performance improvements from the low-learning-rate phase of an ELR cycle than would be expected in supervised learning~\citep{lyle2024normalization}. We therefore propose to perform early-stopping of the cyclic schedule, using a cycle period of 4M optimizer steps and switching to a constant learning rate after ten cycles of re-warming, or 40M frames, for the remainder of training. We observe similar results for other choices of stopping point in Appendix~\ref{sec:appx-rl-es}. 
In Figure~\ref{fig:es-cycling}, we show that this approach results in consistent and dramatic performance improvements in both Rainbow and DQN agents trained in the high-UTD regime on Atari environments. We observe particularly significant improvements in \textit{seaquest} and \textit{krull}, two domains where the cyclic learning rate schedule struggled with instability. The early cycling of the learning rate results in improvements over the constant low learning rate, demonstrating the benefits of \textit{transiently} passing through the high-LR regime (repeatedly). We show in Appendix~\ref{sec:appx-rl-es} that the DQN agent exhibits similar benefits from early-stopping at 8M or 20M frames. 

\section{Conclusion}

This work has shown that re-warming of the effective learning rate can rapidly induce feature-learning, and thus generalization, in a variety of neural network training tasks.
We proposed a simple modification to the Normalize-and-Project method which introduces a non-monotone learning rate schedule and used this approach to facilitate grokking, that is, to overwrite initial features which permitted memorization of the training data with ones that generalize. 
We further showed that the same ELR re-warming techniques designed to facilitate grokking can just as readily be applied to overwrite \textit{learned} features, demonstrating the efficacy of learning rate re-warming at closing the generalization gap induced by warm-starting neural network training. We further observed significant improvements from learning rate re-warming in high-update-to-data reinforcement learning tasks, though we observe much greater instability at high learning rates which requires more aggressive annealing than was required in classification tasks. We conclude that re-warming the effective learning rate is a powerful tool for combating primacy bias in neural networks, but that it must be deployed with care to avoid introducing instabilities which can derail the feature-learning process which it was intended to facilitate.

\newpage
\bibliography{arxiv}
\bibliographystyle{collas2025_conference}

\newpage
\appendix
\onecolumn

\section{Experimental details}

\subsection{Grokking}
\label{sec:exp-details-grokking}
\paragraph{Architecture details} we provide details on the transformer architecture in Table~\ref{tab:transformer}. The architecture consists of an input embedding layer using absolute positional encoding, followed by a single attention block, followed by a two-layer MLP, followed by a decoding layer. LayerNorm layers, when incorporated, are applied to the attention inputs and outputs, and the MLP outputs. We omit bias terms in the network and use RMSNorm by default. 
\begin{table}[]
    \centering
    \begin{tabular}{|c|c|}
    \hline 
        \textbf{Parameter} & \textbf{Value} \\
        \hline \hline
        num layers & 2 \\
         \hline
        num heads &  4 \\
         \hline
        query/key/value dimension&32\\
         \hline
        feed-forward hidden size&512\\
         \hline
          dropout rate & 0\\
           \hline
          relative position embeddings & False\\
           \hline
          absolute position length & 5\\
           \hline
         LayerNorm position & before attention, before MLP, after MLP\\
          \hline
    \end{tabular}
    \caption{Transformer architecture parameters}
    \label{tab:transformer}
\end{table}

\paragraph{Training protocols} In all experiments, we use an Adam optimizer with variable step size and default values for $\beta_1=0.9, \beta_2=0.999$. The dataset is constructed as described in the main body of the paper. The learning rate schedules we consider are a constant learning rate and a cosine warmup learning rate schedule with 1000 warmup steps and a final learning rate of 0.0001. One important note on our division of the dataset into a training and testing subset is that pairs $(x, y)$ and $(y, x)$ are not distinguished, meaning that a network which has learned to treat its inputs symmetrically but has not learned any other generalizing features will attain approximately 20\% accuracy on the test set via this symmetry. This observation casts an intriguing light on the results of Figure~\ref{fig:wd-lr-figure}, where we sometimes see networks drop \textit{below} this 20\% threshold prior to grokking.

\subsection{Warm-Starting}
\label{sec:exp-details-warm}

In our warm-starting experiments we consider two similar network architectures: a ResNet18 architecture~\citep{he2015delving}, and a smaller CNN based on a popular architecture for reinforcement learning~\citep{mnih2015human}. We use an adam optimizer with variable step size and otherwise default hyperparameters. The sub-sampled dataset is selected by specifying the appropriate number of data points in the \texttt{tfds.load} function from the \texttt{tensorflow\_datasets} library.

\textbf{CNN architecture:} for the CNN architecture, we apply three convolutional layers of 5x5, 3x3, and 3x3 kernels in the first, second, and third layers respectively, each with 32 channels. Each convolutional layer is followed by a LayerNorm and then a ReLU nonlinearity. The convolutional outputs are then flattened and fed through a single hidden fully connected layer with a ReLU activation. We apply a linear layer to this output to obtain the prediction logits.

\textbf{ResNet architecture:} the ResNet is a standard ResNet-18~\citep{he2015delving}, with the only notable modification being that we do not include bias terms in the linear transforms, finding that they complicate the estimation of the effective learning rate without improving performance of the model.

\subsection{Reinforcement Learning} 
\label{sec:exp-details-rl}
\paragraph {High UTD regime} We evaluated DQN and Rainbow on 20 games: Asterix,  SpaceInvaders,  Breakout,  Pong,  Qbert,  DemonAttack, Seaquest, WizardOfWor, RoadRunner, BeamRider, Frostbite, CrazyClimber, Assault, Krull, Boxing, Jamesbond, Kangaroo, UpNDown, Gopher, and Hero. We used a reply ratio of 1 (four times the default value). The minimum and maximum learning rate of the cyclic schedule are 1e-6 and 6.25e-5, respectively. The cycle length is 4M environment step. For weight decay experiments, we used a value of 1e-2.

\paragraph {Low UTD regime} We train a Rainbow agent on 57 games in the Arcade Learning Environment benchmark, following the protocol used by \citep{hessel2018rainbow}. We use an adam optimizer with a default learning rate defaulting to 6.25e-5. Cyclic learning rates use this as their maximal value. Training occurs for 200M frames. 

\section{Supplemental Figures}
\label{sec:additional-results}
\subsection{The effect of attention input norm on representation learning}
\label{sec:appx-layernorm}
\begin{figure}
    \centering
    \includegraphics[width=\linewidth]{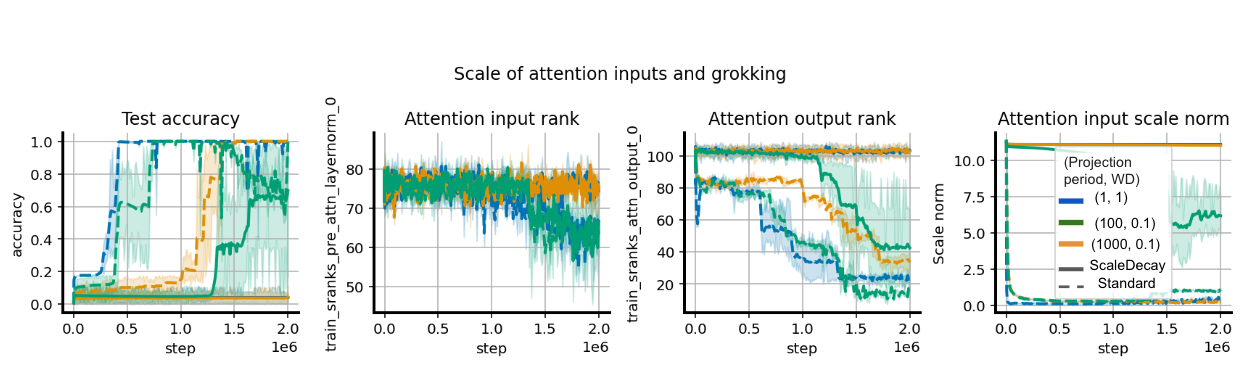}
    \caption{Relationship between attention output rank and grokking in networks trained with LayerNorm and periodic parameter renormalization. Grokking frequently coincides with a reduction in the rank of the attention outputs. Colours correspond to varying weight decay and projection frequency values.}
    \label{fig:layernorm-grokking}
\end{figure}

We observe in the right-hand-side of Figure~\ref{fig:layernorm-grokking} that grokking consistently coincides with a reduction in the rank of the attention head outputs, and that reducing the rank of these outputs in networks with layer normalization requires reducing the norm of either the attention head inputs or the key and query matrices. 

\subsection{Cyclic learning rates and warm-starting}
\label{sec:appx-cyclic-warmstarting}
\begin{figure}
    \centering
    \includegraphics[width=\linewidth]{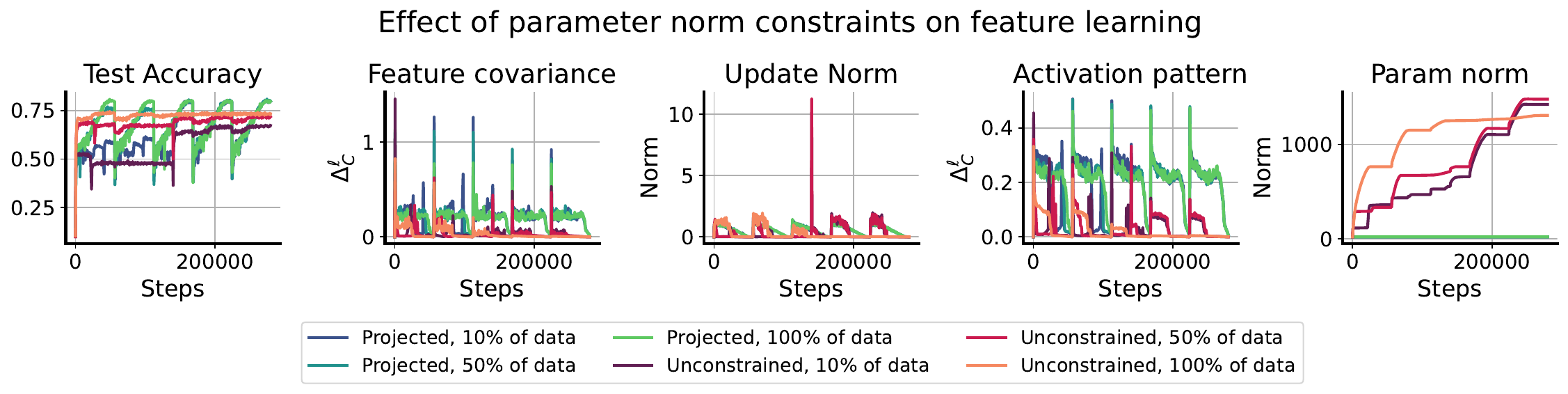}
    \caption{{\color{blue}Replicating the results of Figure~\ref{fig:cyclic-warmstarting}, we see that it is not necessary for the task boundary and ELR re-warming period to align for generalization to benefit from the effects of ELR re-warming.}}
    \label{fig:cyclic-warmstarting-appendix}
\end{figure}

{\color{blue}
While it is important to give time for a cyclic learning rate to complete at least one full cycle after additional data has been added in order to maximally benefit from learning rate re-warming, it is not necessary for this cycle to coincide perfectly with the addition of new data. We replicate the experimental setting of Figure~\ref{fig:cyclic-warmstarting} but now set a cyclic learning rate which re-warms 5 times throughout training, so that the new data is added half-way through the third cycle. We see in Figure~\ref{fig:cyclic-warmstarting-appendix} that this setting also closes the generalization gap.}

\subsection{Ablations on early-stopping of cyclic learning rates in high-UTD RL}
\label{sec:appx-rl-es}
\begin{figure}
    \centering
    \includegraphics[width=\linewidth]{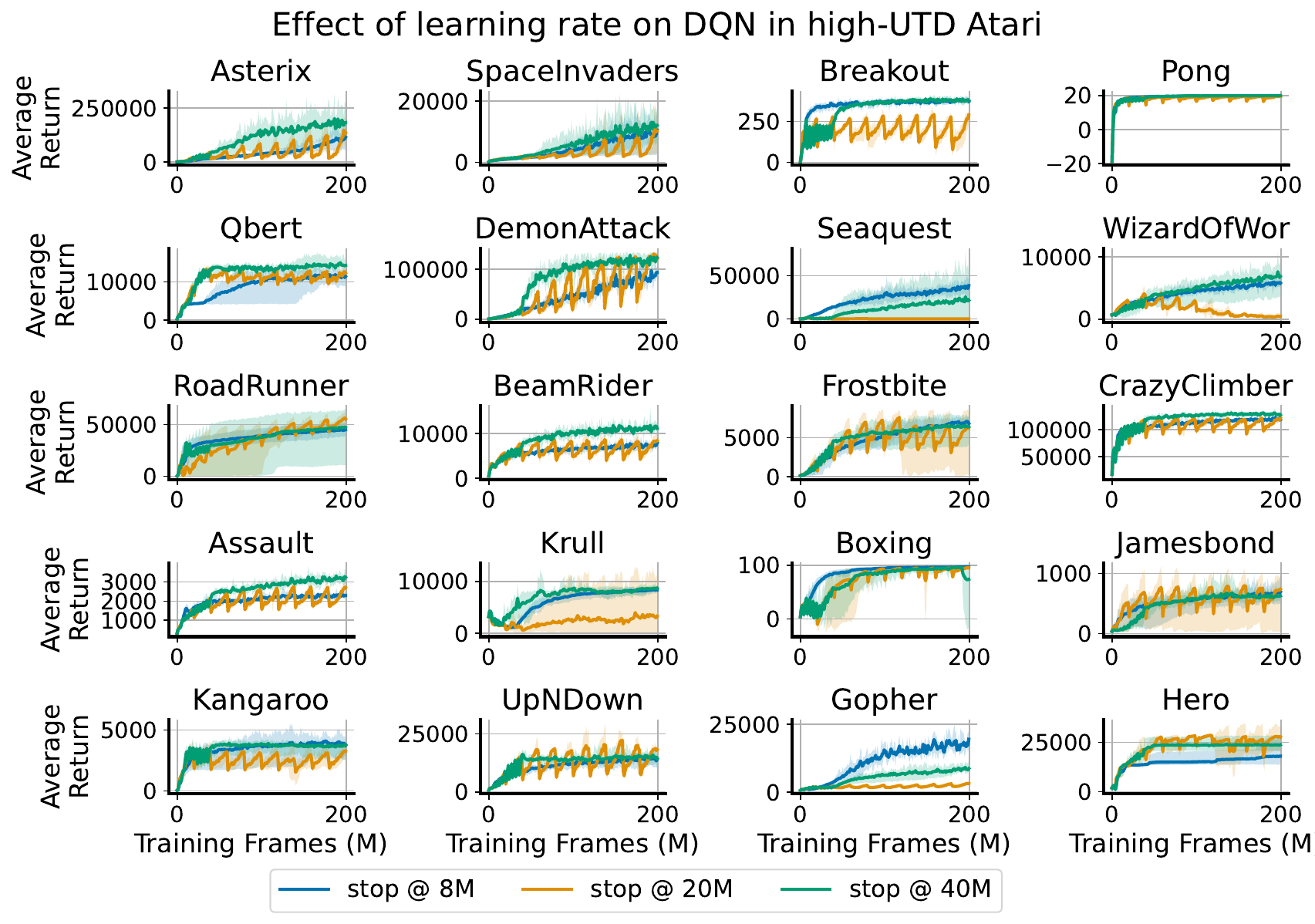}
    \caption{On average, running the cyclic learning rate schedule for longer before annealing results in slightly better performance, but this improvement depends on the game and is not uniform.}
    \label{fig:dqn-es-sweep}
\end{figure}
The choice of when to switch from a cyclic learning rate to a constant low learning rate is somewhat arbitrary. Prior work has observed that a longer low-learning-rate phase of training can be beneficial to agents in the Arcade Learning Environment~\citep{lyle2024normalization}, but did not conduct a thorough sweep. We perform a coarse grid-search over three different stopping points for the learning rate cycling applied to the DQN agent in Figure~\ref{fig:dqn-es-sweep}, observing some variance in outcomes but qualitatively similar effects across all three stopping points when compared to a fixed learning rate. 

\subsection{Full per-game results in low-UTD RL}
\label{sec:full-low-atari}
\begin{figure}
    \centering
    \includegraphics[width=\linewidth]{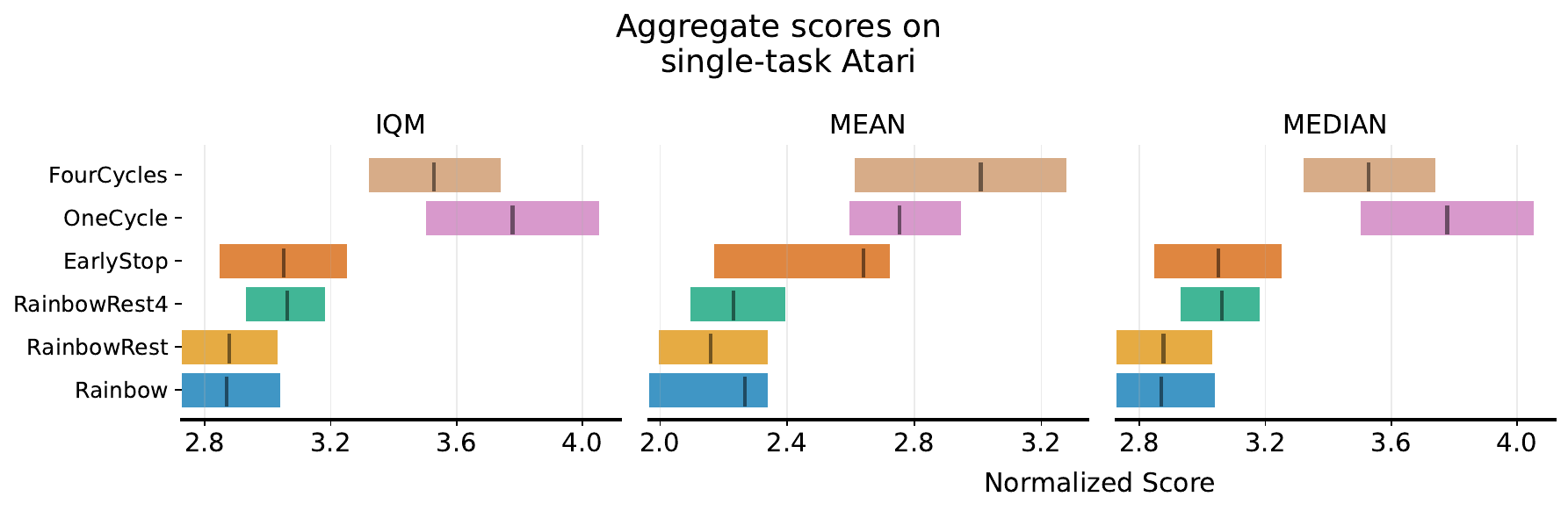}
    \caption{Additional variants on Atari. Early-stopping of the learning rate cycling to a very small learning rate (1e-7) does not improve performance (an exhaustive sweep of terminal values was not feasible due to the computational burden of the benchmark, so it is possible that a higher final value would perform better). Similarly, adding a cyclic learning rate schedule to the reset variant which matches the reset frequency improves performance over the constant learning rate variant but still leaves a large gap with the non-reset variants.}
    \label{fig:iqm-all}
\end{figure}

\begin{figure}
    \centering
    \includegraphics[width=0.98\linewidth]{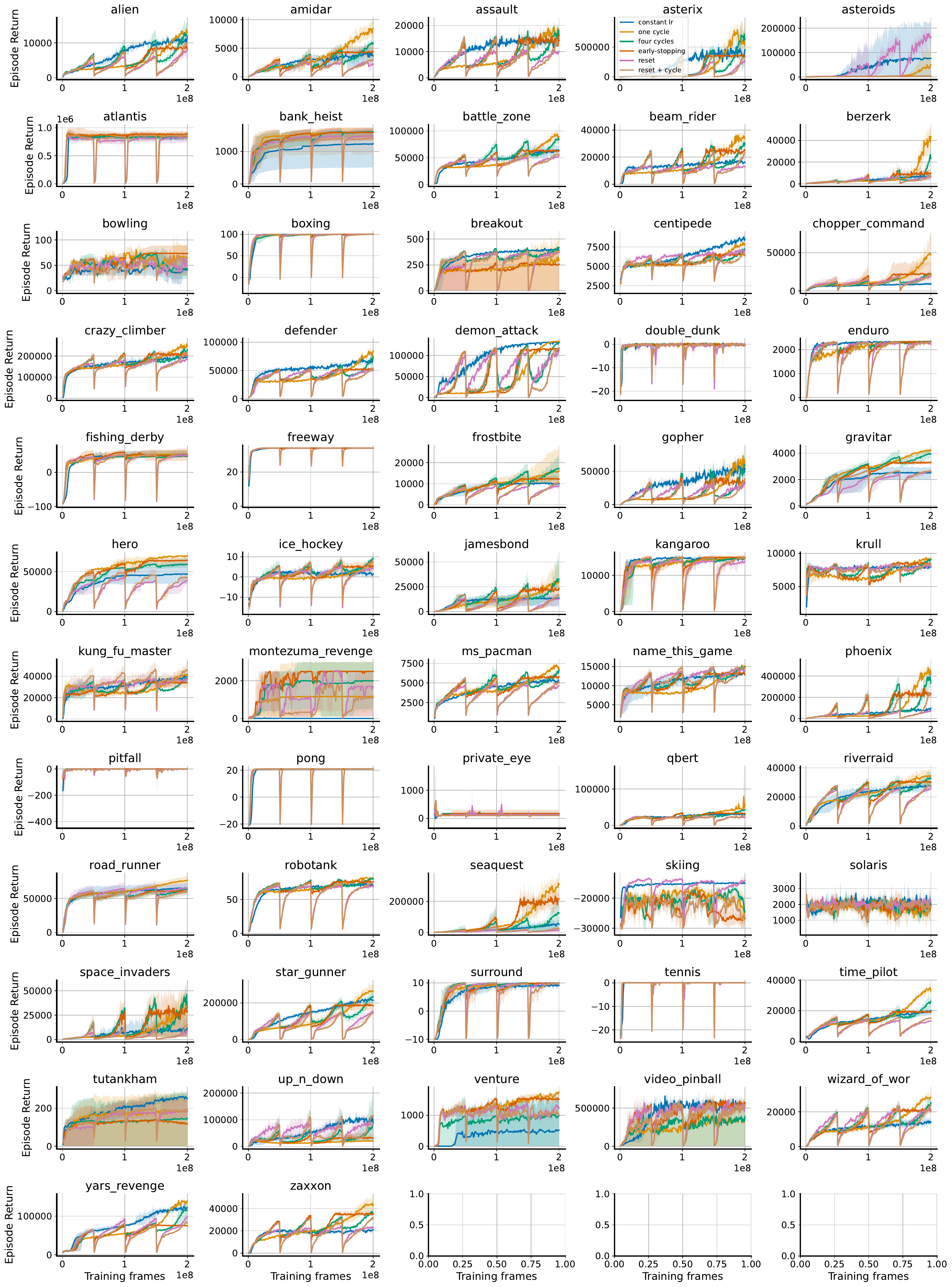}
    \caption{Results for low-UTD rainbow on all atari games.}
    \label{fig:atari-all-online}
\end{figure}

In addition to the variants studied in the main body of the paper, we evaluate two additional variants in Figure~\ref{fig:iqm-all}: an early-stopping variant of the cyclic learning rate schedule which terminates at a final value of 1e-7 after three-quarters of the budget has elapsed, and a variant of the resetting agent that uses a cyclic learning rate schedule that aligns with the reset period. We include per-game results for the Rainbow agent trained in the low-update-to-data regime in Figure~\ref{fig:atari-all-online} for all 57 games.

\subsection{Re-warming ELR above initial value}
\begin{figure}
    \centering
    \includegraphics[height=3.7cm]{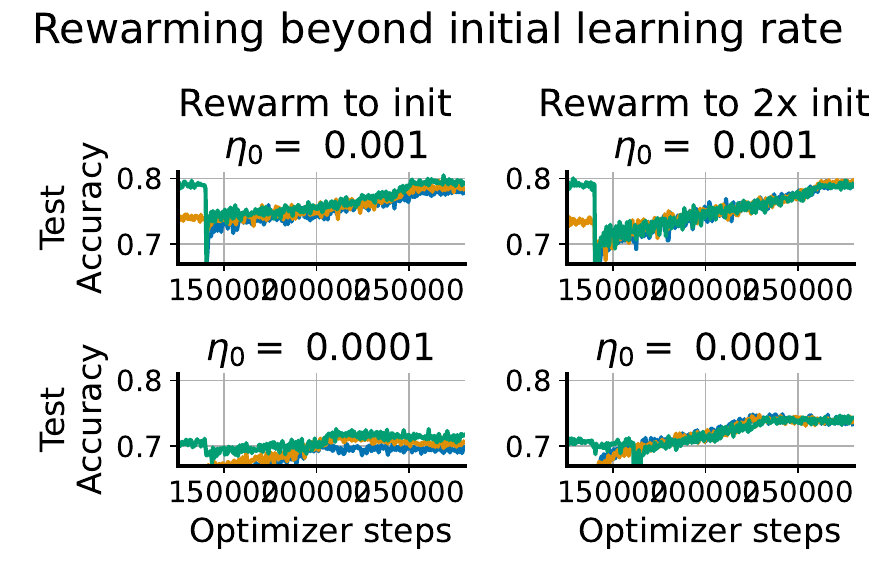}
    \caption{Under some conditions, fully closing the generalization gap requires re-warming the learning rate to a higher value than was used at the start of training -- the small gap between networks trained at different initial data fractions on the left column vanishes on the right after increasing the LR above the value used in the initial training phase.}
    \label{fig:growing-lr}
\end{figure}
In Figure~\ref{fig:growing-lr} we also observe that under some of the per-layer learning rate schedules which do not alone close the warm-starting gap, it is possible to overcome this limitation by re-warming the learning rate to a \textit{larger value than what was used in the initial round of training.} That is, in the setting considered here, it is necessary for the maximal ELR in the second iteration to exceed its maximal value in the first iteration. We demonstrate this finding in the right-hand-side plot of Figure~\ref{fig:per-layer-lrs}, where we see that re-warming the learning rate to a value one order of magnitude above the initial learning rate produces the smallest generalization gaps. The effect of this larger re-warming value on the features can be seen by the sharp increase in rank attained by the larger relative re-warmed values compared to the network trained with identical learning rates.

\clearpage

\section{The effective learning rate and feature evolution}
\label{appx:theory}

It is important to emphasize that our ability to induce feature learning depends not only on increasing the learning rate, but ensuring that the parameter norm does not grow in tandem.

Consider a single-hidden-layer neural network with ReLU activations, and $m$-dimensional input and a one-dimensional output, which can be written in the form
\begin{equation}
    f(x) = w_2^\top \mathrm{Relu}(W_1 x) 
\end{equation}
where $W_1 \in \mathbb{R}^{m \times d}$, $w_2 \in \mathbb{R}^{d}$, $x \in \mathbb{R}^m$ and $\mathrm{Relu}(s)=\max(s,0)$ is the ReLU function (when it is applied to a vector, it is done coordinatewise).

We assume a simplified optimization setting with inputs $\mathbf{X}\in \mathbb{R}^{n \times m}$, where $n$ denotes the number of data points and $m$ the input dimension, so that each row of $\bX$ is of the form $\bx_k$ for some data point $\bx_k \in \mathbb{R}^m$. We will use the notation $\Phi =W_1 \bX$ to refer to pre-ReLU value of the network's hidden features. We assume that gradients can be modeled as random Gaussian perturbations to the parameters. In particular, we assume the gradients on the first layer parameters $W_1$, denoted $\bg \in \mathbb{R}^{m \times d}$, follow the distribution $\bg[i, j] \sim \mathcal{N}(0, \frac{\sigma^2_g}{d})$ so that the expected norm for each row satisfies $\mathbb{E}[\|\bg_i\|] = \sigma^2_g$ (the Gaussian approximation holds when $m$ is large enough and the network is close to initialization, as can be seen, e.g., from the form of the gradient in Equation (2.2) of \citealp{xu2024benign}).
We start with a simple question: how does the angle of an embedding $\Phi_k = W_1 \bx_k$ change as a function of the effective learning rate? 

Consider the embedding of a single input, which we will denote $\Phi_k = W_1\bx_k$. Then, after a single gradient update, we will have $\Phi'_k = (W_1 - \eta \bg)\bx_k$. We assume that $W_1$ has been initialized following the standard fan-in procedure, such that $\|W_1[i, : ]\| \approx 1$, that is, the Euclidean norm of each row of this parameter matrix is approximately 1. Straightforwardly, we have that the expected change in the angle $\theta$ between $\Phi_k$ and $\Phi'_k$ grows as a function of the step size $\eta$. In particular, assuming $\|\eta\bg\| = \sum_{i=1}^m \eta \|\bg_i\| \approx \eta \sigma_g \sqrt{m}$ and that $\langle \bg, W_1[i] \rangle \approx 0$ (reasonable assumptions for sufficiently high-dimensional inputs), we have
\begin{equation}
    \frac{\langle \Phi_k, \Phi'_k\rangle}{\|\Phi_k\|\| \Phi_{k+1}\|} \approx \frac{1}{\sqrt{1 + \eta^2 \sigma^2_g}}
\end{equation}
meaning that the rate at which the \textit{angle} of the embedding changes depends on the learning rate. In a network which is scale-invariant with respect to the parameters of the first layer, i.e. where we have
\begin{equation*}
    f(\bX) = w_2^\top \mathrm{Relu}\left(\frac{W_1\bX}{\|W_1\bX\|}\right) = w_2^\top \tilde{\Phi}
\end{equation*}
then scaling $W_1$ by a scalar $\alpha$ results in the expected rotation of the new features $\tilde{\Phi}'$ obtained after applying a single gradient descent step to $W_1$
\begin{equation*}
     \mathbb{E}[\frac{\langle \tilde{\Phi}_k, \tilde{\Phi}'_k\rangle}{\|\tilde{\Phi}_k\|\| \tilde{\Phi}'_{k}\|} ] \approx \frac{1}{\sqrt{1 + \frac{\eta^2}{\alpha^2} \sigma_g^2}}
\end{equation*}
where the approximation becomes tighter due to the orthogonality of gradients to parameters in scale-invariant functions. 

One can analogously estimate the probability, given a random initialization and a random update, that a currently-activated ReLU unit under some input will be deactivated after a single gradient step. In this case, we have $\Phi_{i, k} \sim \mathcal{N}(0, 1)$ and -- assuming $\|\bx_k\|=1$ -- $\bg_i^\top \bx_k \sim \mathcal{N}(0, \sigma_g^2)$, meaning that the update to $\Phi_{i,k}$ after a gradient step is a Gaussian random variable $\mathbf{z} \sim \mathcal{N}(0, \eta^2 \sigma_g^2)$.  The event we are interested in is thus
\begin{align*}
    P(\Phi_{i, k} - \bz  < 0 | \Phi_{i,k} > 0)  &=  1 - P(\Phi_{i, k} - \bz  > 0 | \Phi_{i,k} > 0) \\
    &= 1 - \bigg(\frac{1}{2} + \frac{1}{\pi} \arctan\frac{1}{\eta\sigma_g} \bigg) \\
    &= \frac{1}{2} - \frac{1}{\pi} \arctan\frac{1}{\eta\sigma_g},
\end{align*}
which holds for $\eta \sigma_g \le 1$ following a derivation from Stackoverflow\footnote{https://math.stackexchange.com/questions/4433691/what-is-pxy0-mid-x0-given-that-x-y-two-different-normal}. So the likelihood of a perturbation changing the activity of a particular unit scales as $\arctan(\frac{1}{\eta})$, assuming our original (not scale-invariant) setup. Incorporating scale-invariance means that now the likelihood of the activity of a particular neuron changing depends on the \textit{effective} learning rate. In particular, supposing now that $W_1$ is scaled by a factor $\alpha$, we obtain
\begin{align*}
    P(\Phi_{i, k} - \tilde{\bz}  < 0 | \Phi_{i,k} > 0)  &=  1 - P(\Phi_{i, k} - \tilde{\bz}  > 0 | \Phi_{i,k} > 0) \; \text{ where } \tilde{\bz} \overset{D}{=} \frac{1}{\alpha}\bz\\
   &= \frac{1}{2} - \frac{1}{\pi} \arctan\frac{\alpha^2}{\eta\sigma_g}
\end{align*}
where we see a term $\alpha^2$ emerge due to the effect of scaling $W_1$ on $\Phi$ as well as its effect on $\bg$. Thus, we see that for large parameter norms, it is necessary to correspondingly increase the learning rate to produce the same effective change in the activation patterns of the representation.

We can also argue why periodically re-warming the learning rate can be useful in non-stationary learning if one uses parameter normalization and some adaptive gradient tuning method. This situation can be modeled as if the parameters $W^\ell$ are constrained to be in the unit ball, and we use an optimizer which reduces the learning rate over the course of training. Running this optimizer on a stationary task usually results in a parameter which (essentially) converged to a boundary of the feasible set, and hence it does not move much no matter what the step size is, so learning-rate re-warming does not really hurt the performance on this task. However, due to the adaptivity of the optimizer, the learning rate is set to some small value by the time this convergence happens. Hence, if the task changes and the optimal feasible parameter for the new task is some other boundary point, due to the very small learning rate, convergence to this point will take very long. On the other hand, an increase in the step size due to re-warming the learning rate can significantly speed up this process.

\end{document}